\documentclass[10pt,twocolumn,letterpaper]{article}

\usepackage[pagenumbers]{iccv} 

\definecolor{iccvblue}{rgb}{0.21,0.49,0.74}
\usepackage[pagebackref,breaklinks,colorlinks,allcolors=iccvblue]{hyperref}

\usepackage{subcaption} 
\usepackage{multirow}
\usepackage{color}
\usepackage{xcolor}
\usepackage{wrapfig}
\usepackage{amsmath}
\usepackage{amsthm}
\usepackage{amssymb}
\usepackage{mathtools}
\usepackage{amssymb}
\usepackage{graphicx}
\usepackage{colortbl}
\usepackage{adjustbox}
\usepackage{cuted}
\usepackage{multicol}
\usepackage{algorithm}
\usepackage{algpseudocode}

\DeclareMathOperator*{\argmax}{arg\,max}
\DeclareMathOperator*{\argmin}{arg\,min}
\DeclareMathOperator{\EX}{\mathbb{E}}
\newcommand{\norm}[1]{\left\lVert#1\right\rVert}

\newtheorem{lemma}{Lemma}
\newtheorem{proposition}{Proposition}

\definecolor{textpurple}{HTML}{730FFF} 

\def\paperID{4326} 
\def\confName{ICCV}
\def\confYear{2025}

\title{SCORE: Soft Label Compression-Centric Dataset Condensation via Coding Rate Optimization}

\author{Bowen Yuan\thanks{The authors contribute equally to the research.}\\
The University of Queensland\\
{\tt\small bowen.yuan@uq.edu.au}
\and
Yuxia Fu\footnotemark[1]\\
The University of Queensland\\
{\tt\small yuxia.fu@uq.edu.au}
\and
Zijian Wang\\
The University of Queensland\\
{\tt\small zijian.wang@uq.edu.au}
\and
Yadan Luo\\
The University of Queensland\\
{\tt\small y.luo@uq.edu.au}
\and
Zi Huang\\
The University of Queensland\\
{\tt\small helen.huang@uq.edu.au}
}

\begin{document}
\maketitle
\begin{abstract}
\vspace{-4.5ex}

Dataset Condensation (DC) aims to obtain a condensed dataset that allows models trained on the condensed dataset to achieve performance comparable to those trained on the full dataset. Recent DC approaches increasingly focus on encoding knowledge into realistic images with soft labeling, for their scalability to ImageNet-scale datasets and strong capability of cross-domain generalization.  However, this strong performance comes at a substantial storage cost which could significantly exceed the storage cost of the original dataset. We argue that the three key properties to alleviate this performance-storage dilemma are informativeness, discriminativeness, and compressibility of the condensed data. Towards this end, this paper proposes a \textbf{S}oft label compression-centric dataset condensation framework using \textbf{CO}ding \textbf{R}at\textbf{E} (SCORE). SCORE formulates dataset condensation as a min-max optimization problem, which aims to balance the three key properties from an information-theoretic perspective. In particular, we theoretically demonstrate that our coding rate-inspired objective function is submodular, and its optimization naturally enforces low-rank structure in the soft label set corresponding to each condensed data. Extensive experiments on large-scale datasets, including ImageNet-1K and Tiny-ImageNet, demonstrate that SCORE outperforms existing methods in most cases. Even with 30$\times$ compression of soft labels, performance decreases by only 5.5\% and 2.7\% for ImageNet-1K with IPC 10 and 50, respectively. Code will be released upon paper acceptance.

\end{abstract}
 \section{Introduction}
\label{sec:intro}


\begin{figure}[h]
        \centering
        \begin{subfigure}[t]{.49\linewidth}
            \centering
            \includegraphics[width=1\linewidth]{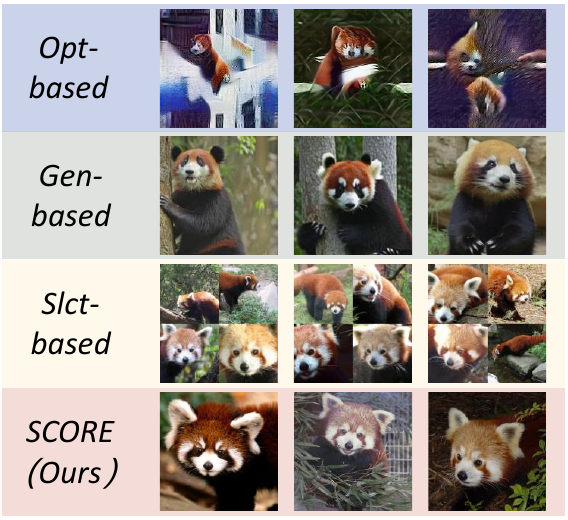}
        \end{subfigure}
        \begin{subfigure}[t]{.49\linewidth}
            \centering
            \includegraphics[width=1\linewidth]{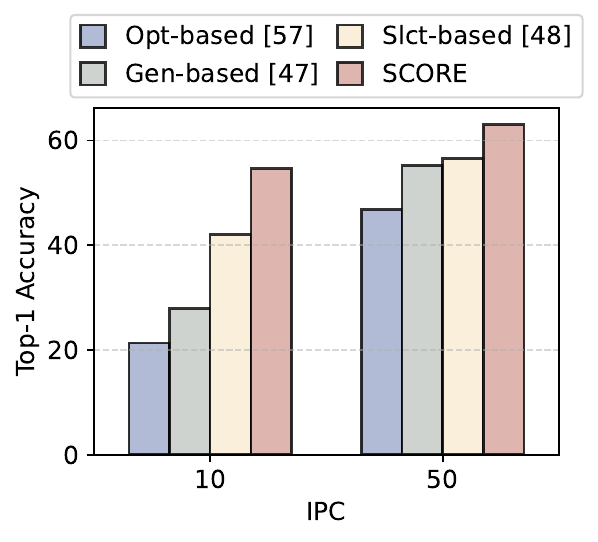}
        \end{subfigure}\vspace{-1ex}
        \caption{We compare the image quality and the corresponding model performance produced by four kinds of DC methods: Optimization-based, Generative model-based, Selection-based and Ours. \textbf{Left:} Visualization of different DC methods.  \textbf{Right:} Performance comparison over different DC methods.}\vspace{-1ex}
        \label{fig:Why Real}
\end{figure}

Dataset condensation (DC) \cite{DD} aims to compress a large-scale training dataset into a significantly smaller synthetic subset that can train downstream models with minimal loss in accuracy. Such condensation not only substantially reduces storage and computational requirements, but also enables flexible and customizable knowledge assembly for various downstream tasks. Prior DC approaches typically formulate dataset condensation as a pixel-level bi-level optimization problem, explicitly aligning training dynamics such as gradient trajectories \cite{trajectory_matching, gradmatch, DATM, TESLA}, data distribution \cite{zhao2023distribution, CAFE, improved_dm}, and parameter evolution \cite{IDC, haba, RTP} between synthetic samples and original sets. While effective on low-resolution benchmarks such as CIFAR \cite{CIFAR} and MNIST \cite{mnist}, these approaches could barely scale to real-world, high-resolution datasets like ImageNet-1K. Key limitations include unaffordable memory consumption that scales \textit{quadratically} with resolution and the limited realism of the yielded synthetic images, which often diverge far from the original data (\textit{1st row} in Figure \ref{fig:Why Real}). The loss of realism arises from the overfitting of sythetic data to specific architectures used during optimization, severely restricting the generalization capacity across diverse downstream models. Consequently, recent studies prioritize enhancing sample \textit{realism} by leveraging generative diffusion models \cite{glad, D4M} (\textit{2nd row}) to synthesize more high-fidelity images or directly selecting informative real samples \cite{rded} (\textit{3rd row}) for ImageNet-scale condensation, leading to improved cross-architecture generalization.

While the visual realism is guaranteed in these approaches, they often sacrifice expressiveness required for effective training, as condensed samples inherently capture reduced intra- and inter-class variations as noted in \cite{rded}. To compensate for this expressiveness gap, recent works commonly employ fast knowledge distillation (FKD) \cite{fkd}, augmenting condensed samples with combined transformations such as \textit{RandomResizeCrop} and \textit{CutMix} \cite{cutmix} and generating the corresponding soft labels from the pretrained teacher models to transfer richer semantic knowledge.

Despite the benefits of soft labels, FKD-based strategies introduce substantial storage overhead, especially when the number of classes is large. For instance, DC approaches equipped with FKD \cite{sre2l, G-VBSM, rded} that condense ImageNet-1K \cite{imagenet} into 200 images per class would require approximately 200 GB for storing soft labels alone, which is almost twice the storage footprint of the \textit{full} ImageNet-1K dataset. Recent efforts \cite{label_worth_thousand_image} have explored the random pruning of soft labels to mitigate this issue; however, blindly discarding label information is suboptimal, inevitably compromising the quality and richness of knowledge captured by the condensed dataset.

Motivated by this storage-performance dilemma, we introduce three fundamental criteria to guarantee realism and expressiveness in dataset condensation simultaneously: (1) \textit{Informativeness} \resizebox{1em}{1em}{\includegraphics{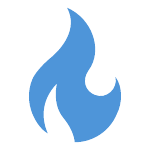}}: selected samples should comprehensively capture the variability inherent in the original dataset; (2) \textit{Discriminativeness} \resizebox{1em}{1em}{\includegraphics{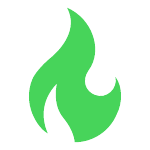}}: condensed data should accurately represent class-specific discriminative information, facilitating downstream classification tasks; and (3) \textit{Compressibility} \resizebox{1em}{1em}{\includegraphics{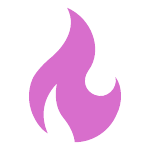}}: augmented soft labels should allow effective compression through rank-preserving techniques such as robust principal component analysis (RPCA), ensuring minimal information loss compared to naive random pruning. Inspired by these, we introduce \textbf{SCORE}, a \textbf{S}oft label compression-centric dataset condensation framework using \textbf{CO}ding \textbf{R}at\textbf{E}. Specifically, we unify these criteria within an information-theoretic min-max optimization framework centered around the concept of coding rate \cite{coding_rate}. The coding rate theoretically serves as a concave surrogate for rank preservation that enables soft label compression. Notably, the coding rate exhibits a desirable submodular property, allowing us to divide and conquer the condensation process into an efficient greedy selection algorithm to balance realism, informativeness, and storage efficiency seamlessly. Extensive evaluations show that SCORE achieves state-of-the-art results across various datasets and network architectures under the same storage budget. Besides, SCORE empirically surpasses existing label compression approaches in DC, achieving improvements of up to 7.9\% on Tiny-ImageNet and 20.3\% on ImageNet-1K. Notably, SCORE reduces storage from 55.9 GB to only 1.9 GB on ImageNet-1K at 50 IPC, while maintaining 95.7\% classification accuracy. The low-rank property induced by the coding rate in the condensed dataset ensures that the effectiveness of our compressed soft labels is agnostic to different compression strategies. By significantly reducing storage overhead while maintaining evaluation performance, SCORE provides a scalable and efficient solution for dataset condensation.

\section{Related Work}
\label{sec:related_work}

\subsection{Dataset Condensation}
Dataset condensation (DC) aims to generate a compact yet informative dataset, enabling models to achieve performance comparable to training on the full dataset. This concept is first introduced by \cite{DD}, which optimizes synthetic data by unrolling model training steps and matching model performance. While groundbreaking, their approach suffers from computationally inefficient bi-level optimization. The following researches propose surrogate objectives to overcome this limitation: distribution matching \cite{zhao2023distribution, CAFE, improved_dm} aligns feature distributions using randomly initialized networks; gradient matching \cite{zhao2021datasetcondensation} and training trajectory \cite{trajectory_matching, DSA, TESLA, DATM} matching ensures model parameter consistency by matching gradients across single or multiple training steps respectively. To generate synthetic images with higher resolution and greater realism, recent works \cite{generative_Cazenavette00EZ23, GAN_dd, D4M, moser2024latent} exploit the power of generative models. 

Most of these methods primarily optimize the images and only assign hard labels to the distilled images. However, soft labels in condensed datasets can even carry more information than synthetic images \cite{label_worth_thousand_image}. Therefore, SRe$^2$L applies data augmentation to synthetic images, generating diverse soft labels for individual samples and achieving substantial performance gains over previous methods on ImageNet-scale datasets \cite{imagenet}. Building on the effectiveness of SRe$^2$L, several dataset distillation methods \cite{G-VBSM, D4M, rded, DWA} have adopted this framework. Although soft labels can yield greater performance gains, their prohibitively large storage requirements contradict the fundamental goal of dataset condensation. In this work, we focus on constructing datasets with informativeness, discriminativeness and compressibility based on the principle of coding rate reduction.

\subsection{Soft Label Compression}

Soft labels improve model performance but demand significantly more storage, especially in SRe$^2$L-like methods. Efficient compression is crucial to maintain their benefits while reducing storage costs. Since soft labels can be represented as matrices, existing matrix compression techniques, such as robust Principal Component Analysis (RPCA) \cite{RPCA} and Singular Value Decomposition (SVD), can decompose them into low-dimensional representations. Some methods exploit the intrinsic redundancy of soft labels, with Zipf’s Label Smoothing \cite{Zipf} using class rankings and FKD \cite{fkd} retaining only top-\( k \) soft scores. Beyond these approaches, Label Pruning for Large-scale Distillation (LPLD) \cite{LPLD} constructs a diverse label pool and randomly samples labels from it to reduce storage costs. However, the random selection process does not guarantee the retention of the most informative labels. In contrast, SCORE applies coding rate reduction, implemented via RPCA, to compress soft labels while preserving the most essential information, which ensures a higher degree of information retention and maintains model performance more effectively.  


\subsection{Coding Rate}

In information theory, the rate distortion measures the minimum average number of bits required to encode data subject to a distortion level $\epsilon$. Due to the difficulty in computing rate distortion, coding rate serves as a practical alternative for its approximation \cite{segcodingrate}. Maximal Coding Rate Reduction (MCR$^2$) \cite{maximal_coding_rate} extends coding rate to facilitate the learning of diverse and discriminative feature representations. MCR$^2$ has been successfully applied to various tasks, including learning natural language representations \cite{macedon}, clustering images \cite{graph_cut_mcr2}, and simplifying self-supervised learning pipelines \cite{ssl_mcr2}. Beyond these applications, coding rate has also been utilized to design interpretable deep networks and Transformer architectures, addressing the issue of catastrophic forgetting \cite{redunet, white_box_transformers, incremental_learning_mcr2}. It has also demonstrated effectiveness in active learning by identifying the most informative samples \cite{kecor}. Unlike previous works, we propose a unified criterion based on coding rate to construct a condensed dataset that meets three properties. 
\section{Critical Challenges in Dataset Condensation}
\label{sec:challenge}

\subsection{Dataset Condensation}
\label{sec:preliminary}

Dataset condensation (DC) addresses the fundamental challenge of condensing large-scale datasets into compact yet representative forms without significant loss of downstream accuracy. Consider an original dataset of $C$ classes $\mathcal{T}={\{{(\mathbf{x}}_{i}, {\mathbf{y}}_{i})\}}^{|\mathcal{T}|}_{i=1}$, where $\mathbf{x}_{i}$ represents the $i$-th image and $\mathbf{y}_{i}$ its corresponding label. Recent DC methods utilize a teacher model $\phi_h(\cdot; \theta_h)$ trained on $\mathcal{T}$ to generate a condensed dataset $\mathcal{S}$. This condensed dataset consists of synthetic images $\Tilde{\mathcal{X}}$ and their associated soft labels. During training process of $K$ iterations, teacher model generates soft labels $\Tilde{\mathcal{Y}} \in \mathbb{R}^{K\times C}$ for each image $\Tilde{\mathbf{x}} \in \Tilde{\mathcal{X}}$ based on various augmentations:  

\begin{equation}
\label{eq:soft_label}
    \Tilde{\mathcal{Y}} =  \{\Tilde{\mathbf{y}}_k = \phi_{h}(A(\Tilde{\mathbf{x}}, k)) \mid \Tilde{\mathbf{x}} \in \mathcal{X}\}^{K}_{k=1},
\end{equation}
where $A(\cdot, \cdot)$ represents the augmentation function. For a downstream student network $\phi(\cdot; \theta)$ trained on the condensed dataset, the objective is to minimize the Kullback-Leibler divergence between the model's predictions and the teacher-generated soft labels across all samples in each iteration $k$:

\begin{equation}
\label{kl divergence function}
\mathcal{L}(\mathcal{S}, \theta) = \sum_{\Tilde{\mathbf{x}} \in \mathcal{X}} \Tilde{\mathbf{y}}_{k} \log \frac{\phi(A(\Tilde{\mathbf{x}}, k))}{\Tilde{\mathbf{y}}_{k}}.
\end{equation}
The keys of dataset condensation are twofold: (1) to achieve significant data compression, ensuring $|\mathcal{S}| \ll |\mathcal{T}|$, and (2) to maintain comparable model performance when training on $\mathcal{S}$ versus $\mathcal{T}$. This can be formulated as a bilevel optimization problem:

\begin{equation}
    \label{bilevel eq}
    \mathcal{S}^* = \underset{\mathcal{S}}{\arg \min}\EX[\mathcal{L}(\mathcal{T}, \theta_{\mathcal{S}})]~ \text{s.t.}~  \theta_{\mathcal{S}} = \underset{\theta}{\argmin}\mathcal{L}(\mathcal{S}, \theta).
\end{equation}




\subsection{Scaling Issues in DC}
\noindent \textbf{Storage Overhead.} While current DC methods with soft labels are effective for large-scale datasets, concerns arise that soft labels overhead grows linearly with the number of classes $C$. Particularly for large-scale datasets such as ImageNet-1K, the storage requirements for soft labels can even exceed those of the condensed images themselves. For example, when obtaining a condensed dataset of ImageNet-1K with 50 image per class (IPC), its corresponding soft labels occupy nearly 56 GB while the images storage is only around 9 GB. Although common compression approaches such as RPCA can resolve the storage delimma, simply compressing soft labels results in severe information loss. As shown in \cref{fig:compare DC comp}, applying the same RPCA compression to soft labels from prior works leads to greater performance decline. When compressing soft labels to 5\% of their original size, prior works suffer accuracy loss of up to 28.7\%, whereas compression on our method maintains better performance, with only 16.1\% performance drop.

\noindent \textbf{Effectiveness of Dataset Synthesis.} When condensing high-resolution datasets, the benefits of synthesizing images are questionable \cite{rded, label_worth_thousand_image}. The contribution of dataset synthesis is limited to evaluation performance. As illustrated in \cref{fig:Why Real}, optimization-based synthetic images fail to preserve realistic visual structures, making them less interpretable. Empirical results shown in bar plots (\cref{fig:Why Real}) demonstrate that DC methods using realistic images outperform those relying on synthetic images. Therefore, our work proposes dataset condensation based on selecting representative images from the original dataset, ensuring the preservation of realistic and crucial visual information. Based upon analysis of DC approaches, we identify three fundamental properties that characterize effective dataset condensation:

\begin{proposition}[Fundamental properties of effective DC]
    \leavevmode
    \begin{enumerate}
        \item[\resizebox{1em}{1em}{\includegraphics{figures/fire_blue.pdf}}] \textbf{\textit{Informativeness}}: The condensed dataset must preserve and efficiently encode the essential information from the original dataset. This includes maximizing the information retention while minimizing the storage footprint.
        \item[\resizebox{1em}{1em}{\includegraphics{figures/fire_green.pdf}}] \textbf{\textit{Discriminativeness}}: To facilitate model classification, the condensed representation should keep clear discriminative boundaries between classes and capture the natural variations present in the original dataset.
        \item[\resizebox{1em}{1em}{\includegraphics{figures/fire_purple.pdf}}] \textbf{\textit{Compressibility}}: The entire condensed representation, including both images and labels, must be storage-efficient. This property has been somewhat overlooked in previous works, which focus primarily on image compression while allowing label storage to grow unconstrained.
    \end{enumerate}
\end{proposition}

\section{Methodology}
\label{sec:method}
\textbf{Overview.}  SCORE employs a greedy selection algorithm to construct a condensed dataset that maintains both realism and essential characteristics of the original data. To reflect three properties, we design a unified goal using coding rate theory \cite{coding_rate}, and the algorithm (shown in \cref{algorithm: coding rate selection}) iteratively builds the condensed dataset by selecting samples from the original dataset using \cref{eq:overall selection}. That is, the algorithm sequentially selects images until the budgets is reached. For each selection, it chooses the candidate sample $\mathbf{x}$ that yields highest \cref{eq:overall selection}, associated with its corresponding soft labels $\mathbf{Y}$ and the current selected image set $\mathcal{X}'$:

\begin{equation}
\label{eq:overall selection}
\begin{split}
    R(\mathbf{x}, \mathbf{Y}|\mathcal{X'}) &=  \underbrace{R_{\operatorname{I}}(f(\{\mathbf{x}\} \cup \mathcal{X'}; \theta_{h}))}_{\text{\textcolor{textpurple}{\resizebox{1.5em}{1.5em}{\includegraphics{figures/fire_blue.pdf}}}}} \\
   & - \alpha \underbrace{R_{\operatorname{D}}(f(\{\mathbf{x}\} \cup \mathcal{X'}; \theta_h))}_{\text{\textcolor{textpurple}{\resizebox{1.5em}{1.5em}{\includegraphics{figures/fire_green.pdf}}}}}
   - \beta \underbrace{R_{\operatorname{C}}(\mathbf{Y})}_{\text{\textcolor{textpurple}{\resizebox{1.5em}{1.5em}{\includegraphics{figures/fire_purple.pdf}}}}},
\end{split}
\end{equation}
where $\alpha$ and $\beta$ are coefficients. The subsequent sections provide detailed analysis of each component in this unified criterion.

\noindent \textbf{Informativeness through Coding Rate Theory.}
In the context of information theory \cite{information_theory}, coding rate function $R_{\operatorname{I}}(\mathbf{Z})$ quantifies the minimum number of bits required to encode a feature space $\mathbf{Z} = [\mathbf{z}_1, ..., \mathbf{z}_n] \in \mathbb{R}^{d \times n}$ within a specified distortion tolerance \cite{rate_distortion_entropy} $\epsilon$:
\begin{equation}
\label{eq:coding rate}
    R_{\operatorname{I}}(\mathbf{Z}) = \frac{1}{2} \log \det \left(\mathbf{I} + \frac{d}{n{\epsilon}^2} \mathbf{Z}\mathbf{Z}^\top\right),
\end{equation}
where $\mathbf{I}$ is the identity matrix and $d$ is the number of dimensions of feature $\mathbf{Z}$. 
In the context of dataset condensation, we aim to use minimum space to reflect maximal information of the original dataset. Therefore, our objective is to \textbf{\textit{maximize}} the coding rate function $R_{\operatorname{I}}(\mathbf{Z})$, given that $\mathbf{Z} = f(\mathcal{X}; \theta_{h})$ represents the features of condensed samples generated by pretrained model.

\noindent \textbf{Enhanced Class Discriminability.} For multi-class classification scenarios, where features naturally cluster into distinct subspaces, we extend the coding rate theory to incorporate class-aware information. Given a dataset with $C$ classes, we define the class-conditional coding rate as:
\begin{equation}
    \mathcal{R}_{\operatorname{D}}(\mathbf{Z}) = \sum_{j=1}^C \frac{\text{tr}(\Pi_j)}{2n} \log \det\left(\mathbf{I} + \frac{d}{\text{tr}(\Pi_j)\epsilon^2}\mathbf{Z}\Pi_j\mathbf{Z}^\top\right),
\end{equation}
where $\Pi_j \in \mathbb{R}^{n \times n}$ is a diagonal matrix and the indices of samples that belong to the $j$-th class are set to be 1, 0 otherwise. This formulation computes the coding rates independently w.r.t. each class, thereby reflecting intra-class sample correlations.  By \textbf{\textit{minimizing}} this term, we promote the sparsification of features across different classes, thus enhancing discriminative power.

\noindent \textbf{Soft Label Compressibility.} We aim to reduce the storage overhead of soft labels by compressing the soft labels, while maintaining their effectiveness. Thus, to maintain the training accuracy with minimal sacrifice, we aim to encourage the soft labels of each image sample low-rank. The soft labels ${\mathbf{Y}} \in \mathbb{R}^{K \times C}$ for an image ${\mathbf{x}}$ are generated by the teacher model under various augmentations. We promote to select samples with low-rank structure in these soft labels by \textbf{\textit{minimizing}} their coding rate:
\begin{equation}
    \begin{split}
        R_{\operatorname{C}}({\mathbf{Y}}) &=  \frac{1}{2} \log \det \left(\mathbf{I} + \frac{d}{n{\epsilon}^2} \mathbf{Y}\mathbf{Y}^\top\right) \\
    & \text{ s.t. } \mathbf{y}_{k} =  \phi_{h}(A(\mathbf{x}, k)).
    \end{split}
\end{equation}

\noindent This term leverages the inherent approximate low-rank nature of soft labels, ensuring intra-class consistency while enabling effective compression.


\section{Theoretical Analysis and Implementations}
\subsection{Why Coding Rate?}
\noindent \textbf{Coding Rate Exhibits Low-Rankness.}
One crucial objective of the work is to compress soft labels to reduce the storage overhead. To achieve effective compression, it is essential that soft labels exhibit low-rankness. If the soft labels of condensed samples possess low-rankness, information loss during compression is minimized.

\begin{lemma}[Coding rate function is a concave surrogate for the rank function \cite{logdet_heuristics}]
\label{lemma:coding rate low rank}
For any matrix $\mathbf{Z} \in \mathbb{R}^{m \times n}$, coding rate function is strictly concave w.r.t. $\mathbf{Z}\mathbf{Z}^\top$ and provides a smooth approximation to the rank function.
\end{lemma}

\noindent In our framework, the coding rate function theoretically conforms to the objective of soft label compression. By minimizing the coding rate of soft labels through our unified objective in \cref{eq:overall selection}, we effectively minimize their rank, thereby enabling effective compression.  The proof is shown in Supplementary.

\noindent \textbf{Realistic Image Selection via Submodularity.}
Submodular functions indicate the information of a set in the context of information theoretics, which have been used in greedy subset selection methods \cite{PRISM, submodular_mutual_coreset}. Submodular functions ensure that greedy algorithms select samples that align with optimal selection objectives \cite{submodular_coreset_activelearning}. A submodular function $f_{\operatorname{sub}}: 2^{\Omega} \rightarrow \mathbb{R}$ on a finite set $\Omega$ satisfies the following condition: for any subsets $\mathcal{M} \subset \mathcal{N} \subseteq \Omega$ and element $\mathbf{x} \in \Omega \setminus{\mathcal{M}}$, the following inequality holds:
\begin{equation}\nonumber
f_{\operatorname{sub}}(\mathcal{M}\cup{\{\mathbf{x}\}}) - f_{\operatorname{sub}}(\mathcal{M}) \geq f_{\operatorname{sub}}(\mathcal{N}\cup{\{\mathbf{x}\}}) - f_{\operatorname{sub}}(\mathcal{N}).
\end{equation}
The marginal gain in the submodular function value when adding an element $x$ to a selected set $\mathcal{X'}$ is given by:
\begin{equation}
    f_{\operatorname{sub}}(\mathbf{x}|\mathcal{X'}) = f_{\operatorname{sub}}(\{\mathbf{x}\} \cup \mathcal{X'}) - f_{\operatorname{sub}}(\mathcal{X'}).
\end{equation}


\begin{lemma}[Coding rate function is submodular]
\label{lemma:coding rate submodular}
Given a set of features $\mathbf{Z}$ from a teacher model $\phi_h(\cdot; \theta_{h})$, coding rate function $R_{\operatorname{I}}(\mathbf{Z})$ satisfies the definition of submodular function.
\end{lemma}
\noindent This submodularity property ensures that our greedy selection algorithm achieves to maximize the coding rate-based objective. Each criterion of dataset condensation properties can be expressed as a conditional submodular function, measuring the marginal gain when adding an image sample to the condensed dataset. 

Overall, by combining the marginal gains, we arrive at unified selection criterion in \cref{eq:overall selection}. The proof is provided in Supplementary and \cref{algorithm: coding rate selection}  aims to select realistic samples with maximal scores w.r.t. \cref{eq:overall selection}.




\subsection{Soft Label Compression}
Since we find the intrinsic connection between coding rate theory and low-rankness, we perform low-rank compression for soft labels. Without loss of generality, we use the robust principle component (RPCA) analysis \cite{RPCA} to compress soft labels. Our strategy is also agnostic to varying compression methods, as demonstrated in \cref{exp:various_compression}. RPCA aims to decompose the soft label matrix $\mathbf{M}$ of a sample into a low-rank component $\mathbf{L}$ and a sparse matrix $\mathbf{S}$, forming a non-convex constraint problem:
\begin{equation}
    \argmin_{\mathbf{L}, \mathbf{S}} rank(\mathbf{L}) + \lambda \norm{\mathbf{S}}_{0}, \text{subject to } \mathbf{M} = \mathbf{L} + \mathbf{S},
\end{equation}
where $\norm{\cdot}_{0}$ is $\ell_0$ norm. We empirically adopt the Principle Component Pursuit (PCP) framework \cite{RPCA, exact_alm, inexact_alm} to compress soft labels.

\begin{algorithm}[!t]
    \caption{Maximal Coding Rate Dataset Condensation}
    \label{algorithm: coding rate selection}
    \begin{algorithmic}
    \Require Real Dataset $\mathcal{T}$; IPC; Teacher Network $\theta_h$
    \Ensure Subset $\mathcal{X}_{\text{select}}$
    
    \State $\mathcal{X'} \gets \emptyset$ \Comment{Initialize selected subset}
    \State $\mathcal{I'} \gets \emptyset$ \Comment{Initialize selected indices}
    \While{$|\mathcal{X'}| < \text{IPC} * C$}
        \State $\mathcal{X}^{c} \gets \text{Batch of candidate images from } \mathcal{T} \setminus \mathcal{X'}$
        \State $j^* \gets \argmax_{j \notin \mathcal{I'}, \mathbf{x}_{j} \in \mathcal{X}^{c}} \mathcal{R}(\mathbf{x}_{j}, \mathbf{Y}_{j} | \mathcal{X'})$ 
        \State $\mathbf{X'} \gets \mathbf{X'} \cup \{\mathbf{x}_{j^*}\}$ 
        \State $\mathcal{I'} \gets \mathcal{I'} \cup \{j^*\}$ 
    \EndWhile\\
    \Return{$\mathcal{X}_{\text{select}}$} 
    \end{algorithmic}
\end{algorithm}

\section{Experiments}
\label{sec:experiments}

\begin{table*}[!htbp]
\centering
\Large 
\resizebox{1\linewidth}{!}{
\renewcommand{\arraystretch}{1.2}
\begin{tabular}{ccccccccccccccc}
\toprule
\multirow{2}{*}{Dataset} & \multirow{2}{*}{IPC} & \multicolumn{2}{c}{MobileNet-V2} & \multicolumn{7}{c}{ResNet-18} & \multicolumn{4}{c}{ResNet-101} \\
\cmidrule(lr){3-4} \cmidrule(lr){5-11} \cmidrule(lr){12-15}
& & EDC & SCORE & D$^4$M \cite{D4M} & SRE$^2$L \cite{sre2l} & G-VBSM \cite{G-VBSM} & RDED  \cite{rded} & LPLD \cite{LPLD} & EDC \cite{edc} & SCORE & SRE$^2$L & RDED & EDC & SCORE \\
\midrule
\multirow{2}{*}{Tiny-ImageNet} 
& 10 & \underline{\textcolor{NavyBlue}{40.6\small${\pm0.6}$}}& \cellcolor{blue!10}\textcolor{Maroon}{\textbf{48.8\small${\pm0.1}$}}& -& 33.1\small${\pm0.4}$& - & 41.9\small${\pm0.2}$ & 35.2\small${\pm0.2}$& \underline{\textcolor{NavyBlue}{51.2\small${\pm0.5}$}}& \cellcolor{blue!10}\textcolor{Maroon}{\textbf{51.3\small${\pm0.2}$}}& 33.9\small${\pm0.6}$ & 22.9\small${\pm3.3}$ & \textcolor{Maroon}{\textbf{51.6\small${\pm0.2}$}} & \cellcolor{blue!10}\underline{\textcolor{NavyBlue}{50.8\small${\pm1.1}$}}\\
& 50 & \underline{\textcolor{NavyBlue}{50.7\small${\pm0.1}$}}& \cellcolor{blue!10}\textcolor{Maroon}{\textbf{57.6\small${\pm0.2}$}}& 46.8\small${\pm0.0}$& 41.1\small${\pm0.4}$ & 47.6\small${\pm0.3}$ & \underline{\textcolor{NavyBlue}{58.2\small${\pm0.1}$}}& 48.8\small${\pm0.4}$ & 57.2\small${\pm0.2}$ & \cellcolor{blue!10}\textcolor{Maroon}{\textbf{58.4\small${\pm0.1}$}}& 42.5\small${\pm0.2}$ & 41.2\small${\pm0.4}$ & \underline{\textcolor{NavyBlue}{58.6\small${\pm0.1}$}}& \cellcolor{blue!10}\textcolor{Maroon}{\textbf{59.6\small${\pm0.6}$}}\\
\midrule
\multirow{2}{*}{ImageNet-1K} 
& 10 & \underline{\textcolor{NavyBlue}{45.0\small${\pm0.2}$}}& \cellcolor{blue!10}\textcolor{Maroon}{\textbf{49.3\small${\pm0.3}$}}& 27.9\small${\pm0.0}$& 21.3\small${\pm0.6}$& 31.4\small${\pm0.5}$& 42.0\small${\pm0.1}$ & 34.6\small${\pm0.9}$ & \underline{\textcolor{NavyBlue}{48.6\small${\pm0.3}$}}& \cellcolor{blue!10}\textcolor{Maroon}{\textbf{54.6\small${\pm0.2}$}}& 30.9\small${\pm0.1}$ & 48.3\small${\pm1.0}$ & \underline{\textcolor{NavyBlue}{51.7\small${\pm0.3}$}} & \cellcolor{blue!10}\textcolor{Maroon}{\textbf{62.5\small${\pm0.3}$}}\\
& 50 & \underline{\textcolor{NavyBlue}{57.8\small${\pm0.1}$}}& \cellcolor{blue!10}\textcolor{Maroon}{\textbf{61.9\small${\pm0.2}$}}& 55.2\small${\pm0.0}$& 46.8\small${\pm0.2}$& 51.8\small${\pm0.4}$& 56.5\small${\pm0.1}$ & 55.4\small${\pm0.3}$ & \underline{\textcolor{NavyBlue}{58.0\small${\pm0.2}$}}& \cellcolor{blue!10}\textcolor{Maroon}{\textbf{63.0\small${\pm0.3}$}}& 60.8\small${\pm0.5}$ & 61.2\small${\pm0.4}$ & \underline{\textcolor{NavyBlue}{64.9\small${\pm0.2}$}}& \cellcolor{blue!10}\textcolor{Maroon}{\textbf{68.5\small${\pm0.4}$}}\\
\bottomrule
\end{tabular}
}
\caption{Performance comparison of models trained on condensed Tiny-ImageNet and ImageNet-1K datasets, generated by using different dataset condensation techniques.  All methods use an identical storage budget, and standard deviations are computed from three independent runs. The best-performing results are highlighted in \textcolor{Maroon}{\textbf{red bold}} text, while the second-best results are marked with \textcolor{NavyBlue}{\underline{blue underlining}}.}
\label{tab:main_results}
\end{table*}

\begin{table*}[!htbp]
\centering
\large
\resizebox{1\linewidth}{!}{
\renewcommand{\arraystretch}{1.2}
\begin{tabular}{cccccccccccccc}
\toprule
\multirow{2}{*}{Dataset} & \multirow{2}{*}{IPC} & \multicolumn{4}{c}{10$\times$} & \multicolumn{4}{c}{20$\times$} & \multicolumn{4}{c}{30$\times$}\\
\cmidrule(lr){3-6} \cmidrule(lr){7-10} \cmidrule(lr){11-14} 
& & SRe$^2$L & RDED & LPLD & SCORE          & SRe$^2$L & RDED & LPLD  & SCORE          & SRe$^2$L & RDED & LPLD  & SCORE \\ 
\midrule
\multirow{2}{*}{Tiny-ImageNet}
& 10 & 30.7\scriptsize${\pm0.2}$$^\ddag$& \underline{\textcolor{NavyBlue}{34.8\scriptsize${\pm0.1}$}}$^\ddag$& 34.2\scriptsize${\pm0.6}$$^\ddag$& \cellcolor{blue!10}\textcolor{Maroon}{\textbf{43.7\scriptsize${\pm0.3}$}}& 28.2\scriptsize${\pm0.0}$$^\ddag$& \underline{\textcolor{NavyBlue}{33.6\scriptsize${\pm0.5}$}}$^\ddag$& 30.3\scriptsize${\pm0.8}$$^\ddag$& \cellcolor{blue!10}\textcolor{Maroon}{\textbf{39.0\scriptsize${\pm0.7}$}}& 24.0\scriptsize${\pm0.2}$$^\ddag$& \underline{\textcolor{NavyBlue}{29.1\scriptsize${\pm0.7}$}}$^\ddag$&  26.6\scriptsize${\pm0.3}$$^\ddag$&  \cellcolor{blue!10}\textcolor{Maroon}{\textbf{35.7\scriptsize${\pm0.5}$}}\\
& 50 & 40.3\scriptsize${\pm0.0}$$^\dag$& \underline{\textcolor{NavyBlue}{53.5\scriptsize${\pm0.2}$}}$^\ddag$& 46.7\scriptsize${\pm0.6}$$^\dag$& \cellcolor{blue!10}\textcolor{Maroon}{\textbf{55.5\scriptsize${\pm0.1}$}}& 39.0\scriptsize${\pm0.0}$$^\dag$& \underline{\textcolor{NavyBlue}{52.3\scriptsize${\pm0.2}$}}$^\ddag$& 44.3\scriptsize${\pm0.5}$$^\dag$& \cellcolor{blue!10}\textcolor{Maroon}{\textbf{53.4\scriptsize${\pm0.3}$}}& 34.6\scriptsize${\pm0.0}$$^\dag$& \textcolor{Maroon}{\textbf{50.2\scriptsize${\pm0.1}$}}$^\ddag$ &  40.2\scriptsize${\pm0.3}$$^\dag$&  \cellcolor{blue!10}\underline{\textcolor{NavyBlue}{49.9\scriptsize${\pm0.2}$}}\\
\midrule
\multirow{2}{*}{ImageNet-1K}
& 10 & 18.9\scriptsize${\pm0.0}$$^\dag$& \underline{\textcolor{NavyBlue}{36.9\scriptsize${\pm0.6}$}}$^\ddag$& 32.7\scriptsize${\pm0.6}$$^\dag$& \cellcolor{blue!10}\textcolor{Maroon}{\textbf{50.9\scriptsize${\pm0.3}$}}& 16.0\scriptsize${\pm0.0}$$^\dag$& \underline{\textcolor{NavyBlue}{33.8\scriptsize${\pm0.1}$}}$^\ddag$& 28.6\scriptsize${\pm0.4}$$^\dag$& \cellcolor{blue!10}\textcolor{Maroon}{\textbf{50.2\scriptsize${\pm0.2}$}}& 14.1\scriptsize${\pm0.0}$$^\dag$& \underline{\textcolor{NavyBlue}{28.8\scriptsize${\pm0.3}$}}$^\ddag$& 23.1\scriptsize${\pm0.1}$$^\dag$& \cellcolor{blue!10}\textcolor{Maroon}{\textbf{49.1\scriptsize${\pm0.3}$}}\\
& 50 & 44.1\scriptsize${\pm0.0}$$^\dag$& \underline{\textcolor{NavyBlue}{54.5\scriptsize${\pm0.2}$}}$^\ddag$& 54.4\scriptsize${\pm0.2}$$^\dag$& \cellcolor{blue!10}\textcolor{Maroon}{\textbf{60.8\scriptsize${\pm0.2}$}}& 41.5\scriptsize${\pm0.0}$$^\dag$& \underline{\textcolor{NavyBlue}{52.5\scriptsize${\pm0.1}$}}$^\ddag$& 51.8\scriptsize${\pm0.2}$$^\dag$& \cellcolor{blue!10}\textcolor{Maroon}{\textbf{60.4\scriptsize${\pm0.5}$}}& 37.2\scriptsize${\pm0.0}$$^\dag$& \underline{\textcolor{NavyBlue}{49.7\scriptsize${\pm0.3}$}}$^\ddag$& 48.6\scriptsize${\pm0.2}$$^\dag$& \cellcolor{blue!10}\textcolor{Maroon}{\textbf{60.3\scriptsize${\pm0.4}$}}\\
\bottomrule
\end{tabular}

}
\caption{
Performance comparison of ResNet-18 trained on condensed Tiny-ImageNet and ImageNet-1K datasets, generated by using different dataset condensation techniques.  We reduce the storage budget of soft labels by 10$\times$, 20$\times$, 30$\times$ for each condensation method. The standard deviations are computed from three independent runs. $^\dag$denotes the reported results, $^\ddag$ denotes the reproduced results. The best-performing results are highlighted in \textcolor{Maroon}{\textbf{red bold}} text, while the second-best results are marked with \textcolor{NavyBlue}{\underline{blue underlining}}.
}
\label{tab:compression results}
\end{table*}

In this section, we present a comprehensive experimental evaluation of our framework. First, we benchmark SCORE against state-of-the-art dataset condensation approaches. Next, we demonstrate the effectiveness of our soft label compression, which significantly reduces storage requirements while maintaining comparable model performance. Finally, we assess the cross-architecture generalization capability of the condensed dataset and investigate the contributions of individual components in our framework.

\subsection{Experimental Setup}
\noindent\textbf{Datasets.} To evaluate the effectiveness of SCORE, we conduct experiments on four large-scale datasets, including ImageNet-1K \cite{imagenet}, Tiny-ImageNet \cite{tinyimagenet}, ImageWoof, and ImageNette \cite{imagewoof&nette}. ImageNet-1K consists of 1,281,167 training images spanning 1,000 categories, with a standard resolution of 256×256. ImageWoof and ImageNette are 10-class subsets of ImageNet-1K, maintaining the same resolution. Tiny-ImageNet is an ImageNet variant where images are resized to 64×64 pixels. Tiny-ImageNet contains 200 classes, with each class having 500 images.

\noindent\textbf{Network Architectures.} Following EDC \cite{edc}, we use the full original dataset (\textit{i.e.}, ImageNet-1k, Tiny-ImageNet, \textit{etc.}) and pretrained ResNet-18 \cite{resnet} to select data and generate the corresponding soft labels. To demonstrate the strong cross-architecture generalization capacity, we leverage the ResNet-18 condensed data to train different mainstream network architectures, including ResNet-18 \cite{resnet}, ResNet-50 \cite{resnet}, ResNet-101 \cite{resnet}, MobileNet-V2 \cite{mobilenetv2}, EfficientNet-B0 \cite{efficientnet}, ConvNext-Tiny \cite{convnext}, ShffleNet-V2 \cite{shufflenet}, as well as Swin Transformer-Tiny \cite{swin}. All student models use the official implementation in torchvision \cite{pytorch}, and will be trained from scratch.



\noindent\textbf{Baselines.} To validate the effectiveness of our proposed methods, we compare SCORE with a comprehensive set of recent state-of-the-art dataset condensation methods. Specifically, we compare with one generative model-based approach (\textit{i.e.}, D$^4$M \cite{D4M}), four optimization-basd approaches (\textit{i.e.}, SRE$^2$L \cite{sre2l}, G-VBSM \cite{G-VBSM}, LPLD \cite{LPLD}, and EDC \cite{edc}), and one selection-based method {\textit{i.e.}, RDED \cite{rded}}. D$^4$M identifies optimal representative image embeddings through clustering techniques and subsequently utilize diffusion models to synthesize condensed images. Optimization-based methods refine synthetic images by aligning the statistical properties on the Batch Normalization layers between the original and condensed datasets. In contrast, RDED identifies and retains the most informative samples to construct a compact yet representative dataset.


\subsection{Experimental Results}
\noindent\textbf{Benchmark Performance Evaluation.} 
We compare SCORE with other dataset condensation approaches. Following previous works \cite{sre2l, D4M, G-VBSM, rded, edc}, we perform all experiments under the same storage budget to ensure a fair comparison. Additionally, to assess performance on smaller datasets, we conduct experiments on ImageWoof and ImageNette. The results for ImageNet-1K and Tiny-ImageNet are presented in \cref{tab:main_results}, while those for ImageWoof and ImageNette are shown in \cref{tab:subset main results under the same storage budget}. Across all datasets, our method consistently outperforms others with ResNet-18 and MobileNet V2 backbones, achieving significantly better results. In addition, with ResNet-101 backbone, it achieves performance on par with competing approaches, further demonstrating the effectiveness of our method. This superior performance stems from our dataset's more realistic images, which offer better generalization capabilities across various network architectures and IPC settings. By optimizing our image selection criteria through maximizing $R_{\operatorname{I}}$ and minimizing $R_{\operatorname{D}}$ (\cref{eq:overall selection}), we ensure our condensed dataset retains both rich information content and high class discriminability, which directly translates to enhanced classification performance.

\noindent\textbf{Performance with Compressed Soft Labels.} To ensure the applicability of dataset condensation on the large-scale dataset, we evaluate how reduced soft label set affects the performances of the model trained on it. 


Following \cite{LPLD}, we compress the soft labels by factors of 10$\times$, 20$\times$ and 30$\times$. Compression is especially crucial for datasets with a larger number of classes and higher IPC values. Therefore, we conduct experiments on Tiny-ImageNet and ImageNet-1K with IPC values of 10 and 50. To assess the effectiveness of compressed datasets, we compare SCORE with LPLD \cite{LPLD}, and we also apply LPLD’s compression method to SRe$^2$L and RDED for further comparison. The experimental results are summarized in \cref{tab:compression results}. Our method consistently outperforms or achieves comparable results across both datasets. On ImageNet-1K at 10 IPC, we surpass the second-best method, RDED, by 14.0\%, 16.4\%, and 20.3\% under 10$\times$, 20$\times$, and 30$\times$ compression, respectively. At 50 IPC, the improvements are 6.3\%, 7.9\%, and 10.6\%. This advantage arises from employing \cref{eq:overall selection}, which strategically selects samples that make soft labels more low-rank, ensuring strong performance even at high compression rates.

\noindent\textbf{Cross-Architecture Generalization.}
To evaluate the generalizability of SCORE, we assess the condensed datasets across a variety of downstream models, including ResNet-18, ResNet-50, ResNet-101, MobileNet-V2, EfficientNet-B0, Swin-Tiny, ConvNext-Tiny, and ShuffleNet-V2. In these experiments, all student models are trained using ResNet-18 as the teacher model. To test the robustness of SCORE across different IPC settings, we consider a wider range of IPC values: 5, 10, 30, and 50. We compare our method with RDED and EDC, and as shown in \cref{tab:cross arch}, SCORE consistently outperforms both baselines across all models and IPC settings, with substantial improvements. These results highlight the versatility of our dataset for models of varying complexities. Notably, for lightweight models such as Swin-Tiny and ShuffleNet-V2 at lower IPC values, SCORE achieves performance improvements of 18.8\%, 12.7\%, 13\%, and 17.4\% over RDED and EDC, respectively. Our realistic images provide superior generalization and robustness across different downstream models while offering greater practical applicability. Furthermore, by leveraging coding rate optimization, SCORE selects samples with higher informativeness and discriminativeness, effectively addressing the limited expressiveness typically associated with real image subsets and significantly enhancing overall performance.

\begin{table}[!htbp]
\centering
\Large 
\resizebox{1\linewidth}{!}{
\renewcommand{\arraystretch}{1.2}
\begin{tabular}{cccccccc}
\toprule
\multirow{2}{*}{Dataset} & \multirow{2}{*}{IPC} & \multicolumn{3}{c}{ResNet-18} & \multicolumn{3}{c}{ResNet-101} \\
\cmidrule(lr){3-5} \cmidrule(lr){6-8}
& & SRE$^2$L & RDED & SCORE & SRE$^2$L & RDED & SCORE \\
\midrule
\multirow{2}{*}{ImageWoof} 
& 10& 30.0$\pm$0.8 & {\underline{\textcolor{NavyBlue}{44.4$\pm$1.8}}}& \cellcolor{blue!10}\textcolor{Maroon}{\textbf{45.1$\pm$1.7}}& 22.3$\pm$0.7 & \underline{\textcolor{NavyBlue}{35.9$\pm$2.1}}&\cellcolor{blue!10}\textcolor{Maroon}{\textbf{42.5$\pm$0.9}}\\
& 50& 41.3$\pm$1.1 & \underline{\textcolor{NavyBlue}{71.7$\pm$0.3}}& \cellcolor{blue!10}\textcolor{Maroon}{\textbf{74.1$\pm$1.2}}& 38.6$\pm$0.5 & \textcolor{Maroon}{\textbf{66.1$\pm$0.3}} & \cellcolor{blue!10}\underline{\textcolor{NavyBlue}{61.0$\pm$0.7}}\\
\midrule
\multirow{2}{*}{ImageNette}
& 10 & 45.7$\pm$1.0 & \underline{\textcolor{NavyBlue}{62.7$\pm$0.8}}& \cellcolor{blue!10}\textcolor{Maroon}{\textbf{71.6$\pm$0.2}}& 36.1$\pm$2.5& \underline{\textcolor{NavyBlue}{53.3$\pm$2.7}}& \cellcolor{blue!10}\textcolor{Maroon}{\textbf{63.9$\pm$1.2}}\\
& 50 & 64.4$\pm$0.7 & \underline{\textcolor{NavyBlue}{84.4$\pm$0.2}}& \cellcolor{blue!10}\textcolor{Maroon}{\textbf{84.5$\pm$0.2}}& 62.8$\pm$0.4& \underline{\textcolor{NavyBlue}{80.9$\pm$0.4}}& \cellcolor{blue!10}\textcolor{Maroon}{\textbf{81.3$\pm$1.4}}\\
\bottomrule
\end{tabular}
}
\caption{Performance comparison of models trained on different condensed ImageWoof and ImageNette datasets.  All methods use an identical storage budget, and standard deviations are computed from three independent runs. The best-performing results are highlighted in \textcolor{Maroon}{\textbf{red bold}} text, while the second-best results are marked with \textcolor{NavyBlue}{\underline{blue underlining}}.}
\label{tab:subset main results under the same storage budget}
\end{table}

\begin{table*}[!htbp]
\centering
\Large
\resizebox{\linewidth}{!}{
\renewcommand{\arraystretch}{1.2}
\begin{tabular}{cccccccccc}
    \toprule
     IPC & Method & ResNet-18 & ResNet-50 & ResNet-101& MobileNet-V2 & EfficientNet-B0 & Swin-Tiny & ConvNext-Tiny & ShuffleNet-V2 \\
    \midrule 
     \multirow{3}{*}{5}
     & RDED \cite{rded} &  30.5&  36.0& 35.5&  21.9&  28.1&  12.4&  25.9&  20.1\\
     & EDC \cite{edc} &  36.6&  40.6& 41.0&  31.8&  39.2&  18.5&  33.9&  15.7\\
     & SCORE & \cellcolor{blue!10}\textbf{39.4}& \cellcolor{blue!10}\textbf{41.8}& \cellcolor{blue!10}\textbf{48.9}& \cellcolor{blue!10}\textbf{36.9}& \cellcolor{blue!10}\textbf{46.1}& \cellcolor{blue!10}\textbf{31.2}& \cellcolor{blue!10}\textbf{42.8}& \cellcolor{blue!10}\textbf{33.1}\\
    \midrule
     \multirow{3}{*}{10}
     & RDED \cite{rded} &  42.0&  46.0& 48.3&  34.4&  42.8&  29.2&  48.3&  19.4\\
     & EDC \cite{edc} &  48.6&  54.1& 51.7&  45.0&  51.1&  38.3&  54.4&  29.8\\
     & SCORE & \cellcolor{blue!10}\textbf{54.6}& \cellcolor{blue!10}\textbf{60.2}& \cellcolor{blue!10}\textbf{62.5}& \cellcolor{blue!10}\textbf{49.3}& \cellcolor{blue!10}\textbf{58.5}& \cellcolor{blue!10}\textbf{49.2}& \cellcolor{blue!10}\textbf{63}& \cellcolor{blue!10}\textbf{37.4}\\
     \midrule 
     \multirow{3}{*}{30}
     & RDED \cite{rded} &  49.9&  59.4& 58.1&  44.9&  54.1&  47.7&  62.1&  23.5\\
     & EDC \cite{edc} &  55.0&  61.5& 60.3&  53.8&  58.4&  59.1&  63.9&  41.1\\
     & SCORE & \cellcolor{blue!10}\textbf{61.3}& \cellcolor{blue!10}\textbf{66.8}& \cellcolor{blue!10}\textbf{67.3}& \cellcolor{blue!10}\textbf{59.6}& \cellcolor{blue!10}\textbf{63.1}& \cellcolor{blue!10}\textbf{64.5}& \cellcolor{blue!10}\textbf{68.1}& \cellcolor{blue!10}\textbf{54.8}\\
     \midrule 
     \multirow{3}{*}{50}
     & RDED \cite{rded} &  56.5&  63.7& 61.2&  53.9&  57.6&  56.9&  65.4&  30.9\\
     & EDC \cite{edc} &  58.0&  64.3& 64.9&  57.8&  60.9&  63.3&  66.6&  45.7\\
     & SCORE & \cellcolor{blue!10}\textbf{63.0}& \cellcolor{blue!10}\textbf{68.4}& \cellcolor{blue!10}\textbf{68.5}& \cellcolor{blue!10}\textbf{61.8}& \cellcolor{blue!10}\textbf{65.9}& \cellcolor{blue!10}\textbf{67.2}& \cellcolor{blue!10}\textbf{69.6}& \cellcolor{blue!10}\textbf{57.8}\\
    \bottomrule
\end{tabular}
}
\caption{Cross-architecture generalization performance on ImageNet-1K.}
\label{tab:cross arch}
\end{table*}

\subsection{Further Analysis}
\begin{figure}[h]
\centering\vspace{-1ex}
\includegraphics[width=0.9\linewidth]{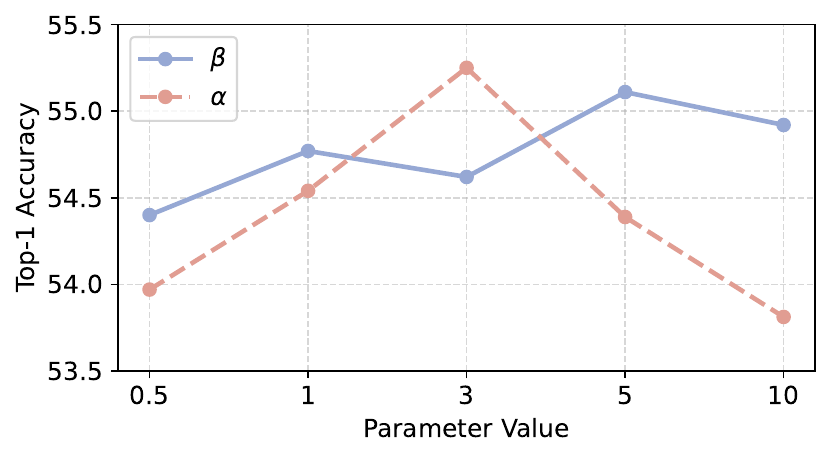}
\caption{Parameter sensitivity analysis of $\alpha$ and $\beta$ in \cref{eq:overall selection}.}\vspace{-2ex}
\label{fig:param sensitive}
\end{figure}

\noindent\textbf{Hyper-parameter Sensitivity Analysis.}
We conduct a sensitivity analysis on the hyper-parameters in \cref{eq:overall selection}, specifically the coefficients $\alpha$ and $\beta$ of the terms $R_{\operatorname{D}}(\mathbf{Z})$ and $R_{\operatorname{C}}({\mathbf{Y}})$. These coefficients control the balance between \textit{Discriminativeness} and \textit{Compressibility}. Theoretically, when $\beta$ is larger, the soft labels of the data tend to experience less information loss during compression. To examine the impact of coefficients individually, we fix $\alpha$ to 5 and vary $\beta$ across the values 0.5, 1, 3, 5, and 10. Additionally, $\beta$ is fixed at 0.5 while $\alpha$ is varied over the same range. Experiments are conducted on ImageNet-1K, with soft label storage compressed to align with the 300-epoch storage budget. Using ResNet-18 as the backbone, we evaluate performance under the 10 IPC setting. As shown in the \cref{fig:param sensitive}, the classification results are sensitive to the choice of $\alpha$ (red dashed line), with the best performance achieved when $\alpha=3$, which is approximately 1.5\% better than the worst result at $\alpha=10$. The effect of different $\beta$ values on the results is relatively steady. Generally, as $\beta$ increases, the performance after compression tends to improve. 



\begin{figure}[h]
\centering
\includegraphics[width=.9\linewidth]{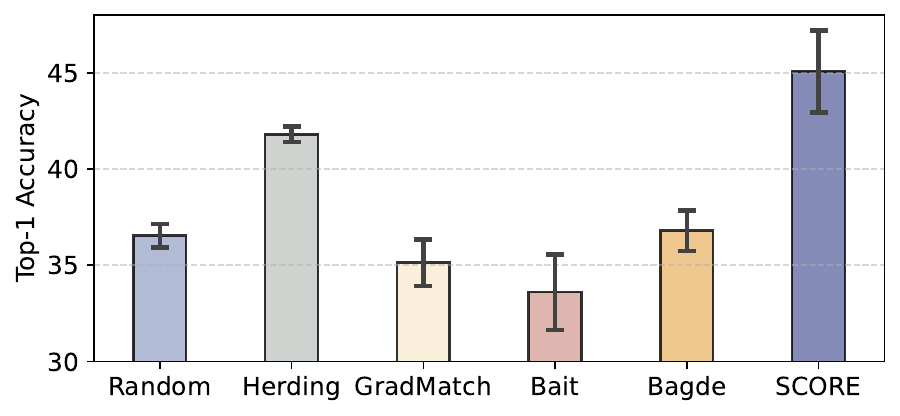}
\caption{The comparison among various coreset selection methods on  ImageWoof with IPC=10.}\vspace{-2ex}
\label{fig:compare Coreset}
\end{figure}

\noindent\textbf{Coreset and Active Learning Method Analysis.}
To assess the effectiveness of selection based on the maximal coding rate criterion, we compare SCORE against various coreset selection strategies on ImageWoof at 10 IPC. The baselines include random selection, Herding \cite{herding}, Bait \cite{bait}, GradMatch \cite{gradmatch}, and BADGE \cite{badge}, an active learning variant that utilizes a pretrained model instead of iterative training. All methods use the same evaluation pipeline with the same storage budget. As shown in \cref{fig:compare Coreset}, SCORE significantly outperforms the alternatives, demonstrating that the dataset constructed using $R_{\operatorname{I}}$ and $R_{\operatorname{D}}$ in the unified criterion \cref{eq:overall selection} preserves more information from the original dataset and exhibits better discriminativeness.

\begin{figure}[h]
\centering
\centering\includegraphics[width=.9\linewidth]{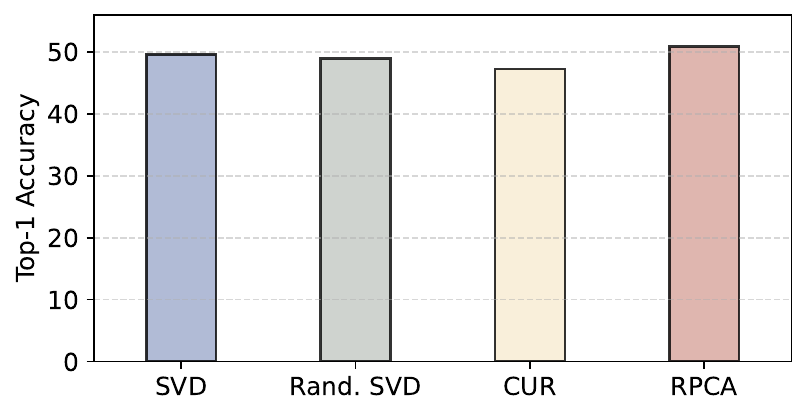}
\caption{The comparison among soft label compression methods on ImageNet-1K with IPC=10.}\vspace{-2ex}
\label{fig:compare comp}
\end{figure}

\noindent\textbf{Sensitivity to Label Compression Choices.}
\label{exp:various_compression}
We evaluate the compressibility of our condensed dataset using various low-rank compression techniques, including SVD \cite{svd}, Randomized SVD \cite{RSVD}, CUR \cite{CUR}, and RPCA \cite{robust_pca}, on ImageNet-1K at 10 IPC. \Cref{fig:compare comp} shows that SCORE exhibits good stability across different compression methods. These results substantiate our original hypothesis: if the soft label set of a condensed dataset inherently exhibits a low-rank structure, it becomes particularly amenable to low-rank compression techniques with minimal information loss. Furthermore, the performance discrepancy observed between the highest-performing method (RPCA) and the lowest-performing alternative is approximately 3.5\%. This relatively minor gap underscores the robustness of our approach in generating soft labels with reliably low-rank characteristics. Considering its superior performance in our comparative analysis, we selected RPCA as the principal compression method for subsequent experiments.

\begin{figure}[h]
\centering\vspace{-1ex}
\includegraphics[width=0.9\linewidth]{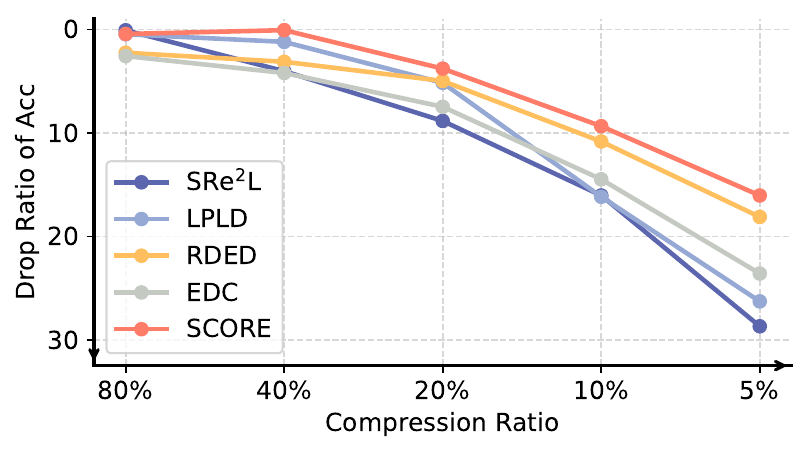}
\caption{The performance drop comparison across different methods.}\vspace{-2ex}
\label{fig:compare DC comp}
\end{figure}

\noindent\textbf{Resilience to Soft Label Compression.}
To demonstrate the advantage of our image selection framework in facilitating soft label compression, we apply RPCA to the soft label set of condensed ImageNet-1K generated by various methods, including SRe$^2$L, RDED, LPLD, EDC, at 10 IPC and train ResNet-18 on the compressed data. Specifically, we establish 300 epochs as the baseline and progressively compress these datasets to 5\%, 10\%, 20\%, 40\%, and 80\% of the original storage. All methods are evaluated under the same training pipeline. As shown in \cref{fig:compare DC comp}, besides demonstrating superior performance before compression, SCORE consistently experiences a smaller performance degradation across all compression rates compared to other methods. This observation suggests that optimizing the \textit{Compressibility} term, $R_{\operatorname{C}}({\mathbf{Y}})$, significantly contributes to stabilizing the variability of soft labels derived from data augmentation. Thus, our selection criterion produces datasets with enhanced resilience to information loss during compression, highlighting the strong potential of our approach to significantly reduce storage requirements while maintaining robust model performance.

\section{Conclusion}

\label{sec:conclusion}
In this study, we addressed the soft label storage overhead challenge of dataset condensation for large scale datasets. We identified three fundamental properties essential for dataset condensation: informativeness, discriminativeness and compressibility. Based on these principles, we proposed SCORE, a dataset condensation method that reduces soft label overhead while maintaining performance. We designed a unified criterion using code rate theory to select realistic images for constructing condensed dataset. SCORE employs RPCA for low-rank soft label compression, while minimizing information loss through our dataset condensation objectives. Experimental results demonstrate that our condensed datasets consistently outperform those from existing methods in downstream model evaluation. The unified selection criterion enables the condensed datasets with informative samples, higher soft label compression ratios and minimal information loss, thereby enhancing the practicality of dataset condensation.
\newpage


{
    \small
    \bibliographystyle{ieeenat_fullname}

}

\clearpage
\onecolumn
\setcounter{page}{1}
\maketitle
\noindent This supplementary material provides additional descriptions of proposed dataset condensation method SCORE, including theoretical proofs and empirical details. Visualizations of condensed datasets are demonstrated to enhance understanding of the proposed method. 
\begin{itemize}
    \item \textbf{\cref{proof:rank_lemma}:} Proof of Coding Rate for Rank Approximation.
    \item \textbf{\cref{proof:submodular_lemma}:} Proof of Coding Rate Submodularity.
    \item \textbf{\cref{proof:criterion_derive}:} Unified Criterion Derivation.
    \item \textbf{\cref{sec:param setting}:} Hyper-Parameter Settings.
    \item \textbf{\cref{sec:visualization}:} Visualizations.
\end{itemize}









\section{Proof}
\subsection{Proof of Lemma 1}
\label{proof:rank_lemma}
Log-det is justified as a smooth approximation for the rank function \cite{logdet_heuristics}. For any matrix $\mathbf{Z} \in \mathbb{R}^{m \times n}$, coding rate function is strictly concave w.r.t. $\mathbf{Z}\mathbf{Z}^\top$ and provides a smooth approximation to the rank function.

\begin{proof}

For all vectors $\mathbf{x}$, given a coefficient $\lambda$ and feature matrix $\mathbf{Z}$, we have
\begin{equation}
\begin{split}
    \mathbf{x}^{\top}(\mathbf{I} + \lambda \mathbf{Z}\mathbf{Z}^{\top})\mathbf{x} &= \mathbf{x}^{\top}\mathbf{I}\mathbf{x} + \lambda\mathbf{x}^{\top}\mathbf{Z}\mathbf{Z}^{\top}\mathbf{x} \\
    &= \mathbf{x}^{\top}\mathbf{x} + \lambda\mathbf{x}^{\top}\mathbf{Z}^{\top}\mathbf{Z}\mathbf{x} \\
    &= {\lVert{\mathbf{x}}\rVert}^2 + \lambda{\lVert{\mathbf{Z}^{\top}\mathbf{x}}\rVert}^{2} \\
    &\geq 0.
\end{split}
\end{equation}

Therefore, the term $\mathbf{I} + \lambda\mathbf{Z}\mathbf{Z}^{\top}$ is positive semidefinite. Now let $\mathbf{Z}\mathbf{Z}^{\top} = \mathbf{M}$. Assume $\mathbf{M}_1$ and $\mathbf{M}_2$ be positive semidefinite matrices, and let $0 \leq \alpha \leq 1$. Define:
\begin{align*} 
X_1 &= \mathbf{I} + \lambda\mathbf{M}_1, \\ 
X_2 &= \mathbf{I} + \lambda\mathbf{M}_2.
\end{align*}

For any convex combination:
\begin{align}
\alpha X_1 + (1-\alpha)X_2 &= \alpha(\mathbf{I} + \lambda\mathbf{M}_1) + (1-\alpha)(\mathbf{I} + \lambda\mathbf{M}_2) \\
&= \mathbf{I} + \lambda(\alpha\mathbf{M}_1 + (1-\alpha)\mathbf{M}_2).
\end{align}

$\log\det(X)$ is concave on the set of positive definite matrices. Therefore:
\begin{equation}
\log\det(\alpha X_1 + (1-\alpha)X_2) \geq \alpha\log\det(X_1) + (1-\alpha)\log\det(X_2)
\end{equation}

Substituting to the definitions:
\begin{align}
\log\det(\mathbf{I} + \lambda(\alpha\mathbf{M}_1 + (1-\alpha)\mathbf{M}_2)) \geq \alpha\log\det(\mathbf{I} + \lambda\mathbf{M}_1) + (1-\alpha)\log\det(\mathbf{I} + \lambda\mathbf{M}_2).
\end{align}

This confirms that $f(\mathbf{M}) = \log\det(\mathbf{I} + \lambda \mathbf{M})$ is concave in $\mathbf{M}$.

Let $\mathbf{M}$ have eigenvalues $\mu_1, \mu_2, \ldots, \mu_n$ (all non-negative since $\mathbf{M}$ is positive semidefinite).

Then $\mathbf{I} + \lambda \mathbf{M}$ has eigenvalues $1 + \lambda\mu_1, 1 + \lambda\mu_2, \ldots, 1 + \lambda\mu_n$.

The determinant is the product of eigenvalues, so:
\begin{equation}
\det(\mathbf{I} + \lambda \mathbf{M}) = \prod_{i=1}^n (1 + \lambda\mu_i)
\end{equation}

Taking the logarithm:
\begin{equation}
\log\det(\mathbf{I} + \lambda \mathbf{M}) = \sum_{i=1}^n \log(1 + \lambda\mu_i)
\end{equation}

Now, as $\lambda \to \infty$:
\begin{itemize}
\item For $\mu_i > 0$: $\log(1 + \lambda\mu_i) \approx \log(\lambda\mu_i) = \log(\lambda) + \log(\mu_i)$
\item For $\mu_i = 0$: $\log(1 + \lambda\mu_i) = \log(1) = 0$
\end{itemize}

Let $r = rank(\mathbf{M})$, which is the number of positive eigenvalues. For a non-zero $\lambda$, we have:
\begin{align}
\log\det(\mathbf{I} + \lambda \mathbf{M}) &\approx \sum_{i: \mu_i > 0} \log(\lambda\mu_i) \\
&= \sum_{i: \mu_i > 0} \log(\lambda) + \sum_{i: \mu_i > 0} \log(\mu_i) \\
&= r\log(\lambda) + \sum_{i: \mu_i > 0} \log(\mu_i)
\end{align}

Thus, $\log\det(\mathbf{I} + \lambda \mathbf{M}) \approx r\log(\lambda) + c$, where $c = \sum_{i: \mu_i > 0} \log(\mu_i)$ is a constant that depends on the non-zero eigenvalues of $\mathbf{M}$.

This shows that for large $\lambda$, minimizing $\log\det(\mathbf{I} + \lambda \mathbf{M})$ is approximately equivalent to minimizing the rank of $\mathbf{M}$, up to scaling by $\log(\lambda)$ and an additive constant.

\end{proof}

\subsection{Proof of Lemma 2}
\label{proof:submodular_lemma}
Given a set of features $\mathbf{Z}$, coding rate function $R_{\operatorname{I}}(\mathbf{Z}) = \log\det(\mathbf{I} + \lambda\mathbf{Z}\mathbf{Z}^{\top})$ satisfies the definition of submodular function. To prove submodularity, $R_{\operatorname{IC}}$ should satisfy diminishing returns property: for any sets $\mathcal{A} \subseteq \mathcal{B}$ and any element $i \notin \mathcal{B}$, if a function $f$ is submodular, we have $f(\mathcal{A}\cup\{i\}) - f(\mathcal{A}) \geq f(\mathcal{B}\cup\{i\}) - f(\mathcal{B})$. Let $\mathcal{A} \subseteq \mathcal{B}$ be two sets of indices for $\mathbf{Z}$, we need to show:
\begin{equation}
    R_{\operatorname{I}}(\mathbf{Z}_{\mathcal{A} \cup \{i\}}) - R_{\operatorname{I}}(\mathbf{Z}_{\mathcal{A}}) \geq R_{\operatorname{I}}(\mathbf{Z}_{\mathcal{B} \cup \{i\}}) - R_{\operatorname{I}}(\mathbf{Z}_{\mathcal{B}}).
\end{equation}


\begin{proof}

Given a set $\mathcal{S}$, we can rewrite the term using matrix determinant lemma:
\begin{equation}
\begin{split}
    R_{\operatorname{IC}}(\mathbf{Z}_{\mathcal{S}} \cup \{i\}) - R_{\operatorname{IC}}(\mathbf{Z}_{\mathcal{s}}) &= \log{\det(\mathbf{I} + \lambda\mathbf{Z}_{\mathcal{S} \cup \{i\}}\mathbf{Z}^{\top}_{\mathcal{S}\cup{\{i\}}})} - \log \det(\mathbf{I} + \lambda\mathbf{Z}_{\mathcal{S}}\mathbf{Z}^{\top}_{\mathcal{S}}) \\
    &= \log{\frac{\det(\mathbf{I} + \lambda\mathbf{Z}_{\mathcal{S} \cup \{i\}}\mathbf{Z}^{\top}_{\mathcal{S}\cup{\{i\}}})}{\det(\mathbf{I} + \lambda\mathbf{Z}_{\mathcal{S}}\mathbf{Z}^{\top}_{\mathcal{S}})}} \\
    &= \log(1 + \lambda\mathbf{z}_{i}^{\top}(\mathbf{I} + \lambda \mathbf{Z}_{\mathcal{S}}\mathbf{Z}_{\mathcal{S}}^{\top})^{-1}\mathbf{z}_{i}). 
\end{split}
\end{equation}

\noindent Since $\mathcal{A} \subseteq \mathcal{B}$, thus $\mathbf{Z}_{\mathcal{A}}\mathbf{Z}_{\mathcal{A}}^{\top}$ is a principle submatrix of $\mathbf{Z}_{\mathcal{B}}\mathbf{Z}_{\mathcal{B}}^{\top}$, thus we have:
\begin{equation}
    \mathbf{Z}_{\mathcal{B}}\mathbf{Z}_{\mathcal{B}}^{\top} - \mathbf{Z}_{\mathcal{A}}\mathbf{Z}_{\mathcal{A}}^{\top} \succeq 0,
\end{equation}
where $\succeq$ denotes the Löwner partial order. Using matrix inversion lemma, we can show that:

\begin{align*}
    {(\mathbf{I} + \lambda \mathbf{Z}_{\mathcal{A}}\mathbf{Z}_{\mathcal{A}}^{\top})}^{-1} - {(\mathbf{I} + \lambda \mathbf{Z}_{\mathcal{B}}\mathbf{Z}_{\mathcal{B}}^{\top})}^{-1} &\succeq 0 \\
    \mathbf{z}_{i}^{\top}[{(\mathbf{I} + \lambda \mathbf{Z}_{\mathcal{A}}\mathbf{Z}_{\mathcal{A}}^{\top})}^{-1} - {(\mathbf{I} + \lambda \mathbf{Z}_{\mathcal{B}}\mathbf{Z}_{\mathcal{B}}^{\top})}^{-1}]\mathbf{z}_{i} &\geq 0.
\end{align*}

Therefore, $R_{\operatorname{I}}(\mathbf{Z}_{\mathcal{s}} \cup \{i\}) - R_{\operatorname{I}}(\mathbf{Z}_{\mathcal{s}})$ is monotonically decreasing, which proves that: given $\mathcal{A} \subseteq \mathcal{B}$, the inequality holds:
\begin{equation}
    R_{\operatorname{I}}(\mathbf{Z}_{\mathcal{A}} \cup \{i\}) - R_{\operatorname{I}}(\mathbf{Z}_{\mathcal{A}}) \geq R_{\operatorname{I}}(\mathbf{Z}_{\mathcal{B}} \cup \{i\}) - R_{\operatorname{I}}(\mathbf{Z}_{\mathcal{B}}),
\end{equation}
and $R_{\operatorname{I}}$ is a submodular function.
\end{proof}

\section{Derivation for Unified Criterion}
\label{proof:criterion_derive}
In the image selection process, we aim to select samples that maximize the marginal gain of the unified criterion in each round. For a selected set $\mathcal{S}$, when a function $f_{\operatorname{sub}}$ is submodular, the marginal gain of adding element $i$ to $\mathcal{S}$ is defined as:

\begin{equation}
    f_{\operatorname{sub}}(\{i\}|\mathcal{S}) = f_{\operatorname{sub}}(\mathcal{S}\cup\{i\}) - f_{\operatorname{sub}}(\mathcal{S}).
\end{equation}

\noindent Based on this submodularity property, given that certain terms remain constant during a single selection iteration, the optimal image $\mathbf{x}^*$ can be determined as follows:

\begin{equation}
\begin{split}
    x^{*} &= \underset{x}{\arg \max} R(\mathbf{x}, \mathbf{Y}|\mathcal{X}') \\
    &= \underset{x}{\arg \max} R_{\operatorname{I}}(f(\{\mathbf{x}\} \cup \mathcal{X}'; \theta_{h})) - R_{\operatorname{I}}(f(\mathcal{X}'; \theta_{h})) \\
    &\quad - \alpha R_{\operatorname{D}}(f(\{\mathbf{x}\} \cup \mathcal{X}'; \theta_{h})) + \alpha R_{\operatorname{D}}(f(\mathcal{X}'; \theta_{h})) - \beta R_{\operatorname{C}}(\mathbf{Y}) \\
    &= \underset{x}{\arg \max} R_{\operatorname{I}}(f(\{\mathbf{x}\} \cup \mathcal{X}'; \theta_{h}))  - \alpha R_{\operatorname{D}}(f(\{\mathbf{x}\} \cup \mathcal{X}'; \theta_{h})) - \beta R_{\operatorname{C}}(\mathbf{Y}).
\end{split}
\end{equation}


\section{Hyper-Parameter Settings}
\label{sec:param setting}
All experiments utilize the AdamW optimizer for model training and apply \textit{RandomResizeCrop}, \textit{RandomHorizontalFlip}, and \textit{cutmix} as data augmentations. The specific hyperparameter settings for all datasets are detailed in \cref{tab:parameters value}. These include the values of $\alpha$ and $\beta$ used for image selection, the minimum and maximum scales for the \textit{RandomResizeCrop} augmentation, as well as other network parameters such as learning rate, weight decay, temperature $T$, batch size, and the number of training epochs.

\begin{table}[ht]
\caption{Hyper-parameter on the ImageNet-1K, Tiny-ImageNet, ImageWoof and ImageNette.}
\centering
  \begin{tabular}{c|cc|cc|cc|cc}
   \toprule
 &\multicolumn{2}{c|}{ImageNet-1K}&\multicolumn{2}{c|}{Tiny-ImageNet}&\multicolumn{2}{c|}{ImageWoof}&\multicolumn{2}{c}{ImageNette}\\
 \cmidrule(lr){2-3} \cmidrule(lr){4-5} \cmidrule(lr){6-7} \cmidrule(lr){8-9}
&\multicolumn{2}{c|}{IPC}&\multicolumn{2}{c|}{IPC}&\multicolumn{2}{c|}{IPC}&\multicolumn{2}{c}{IPC}\\

Hyper-parameter&10&50&10&50&10&50&10&50\\
\midrule

$\alpha$ &5&5&5&10&5&5&0.1&0.1\\
$\beta$ &1&1&0.001&0.001&0.1&0.1&0.1&0.1\\
Min scale of Aug&0.4&0.4&0.3&0.4&0.3&0.3&0.2&0.2\\
Max scale of Aug&1&1&1&1&1&1&1&1\\
Learning Rate& 0.001& 0.001& 0.001& 0.001& 0.001& 0.001&0.0005&0.0005\\
Weight Decay&0.01&0.01&0.01&0.01&0.01&0.01&0.01&0.01\\
$T$&20&20&20&20&20&20&10&10\\
Batch Size&64&64&50&50&64&64&10&10\\

\bottomrule
     \end{tabular}
 \label{tab:parameters value}
\end{table}

\section{Visualization}
\label{sec:visualization}
\begin{figure*}[!htbp]
  \centering\vspace{-5ex}
    \includegraphics[width=\linewidth]{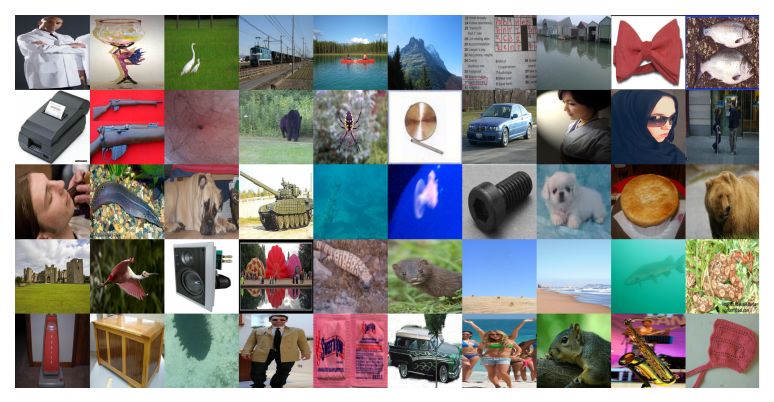}\vspace{-3ex}
  \caption{visualization of the condensed ImageNet-1k.}
  \label{fig:vis_1k}
\end{figure*}

\begin{figure*}[htbp]
  \centering\vspace{-2ex}
    \includegraphics[width=\linewidth]{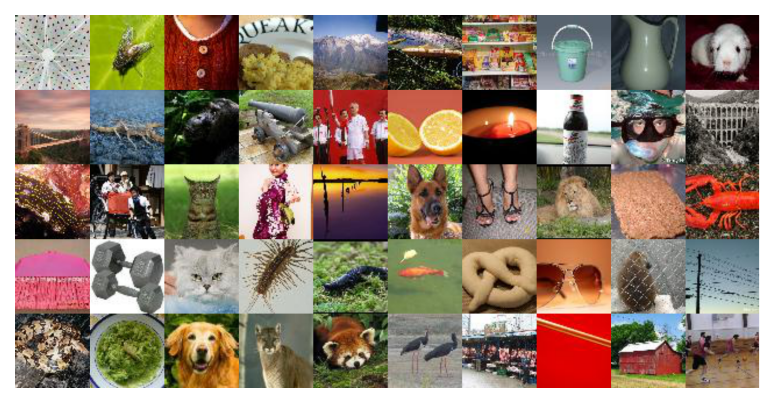}\vspace{-3ex}
  \caption{visualization of the condensed Tiny-ImageNet.}
  \label{fig:vis_tiny}
\end{figure*}

\begin{figure}[htbp]
  \centering\vspace{-2ex}
    \includegraphics[width=\linewidth]{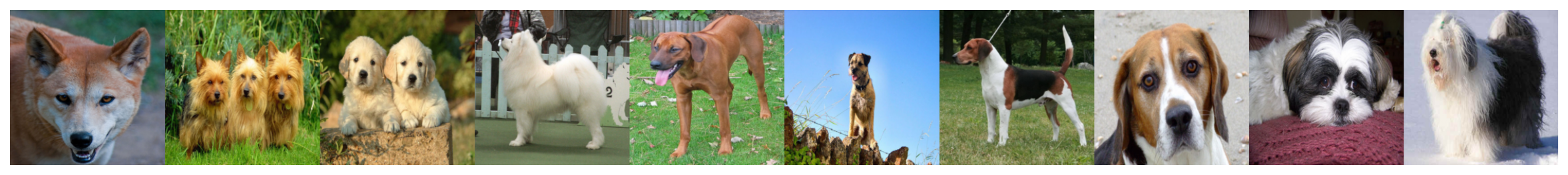}\vspace{-3ex}
  \caption{visualization of the condensed ImageWoof.}
  \label{fig:vis_woof}
\end{figure}

\begin{figure}[!htbp]
  \centering\vspace{-2ex}
    \includegraphics[width=\linewidth]{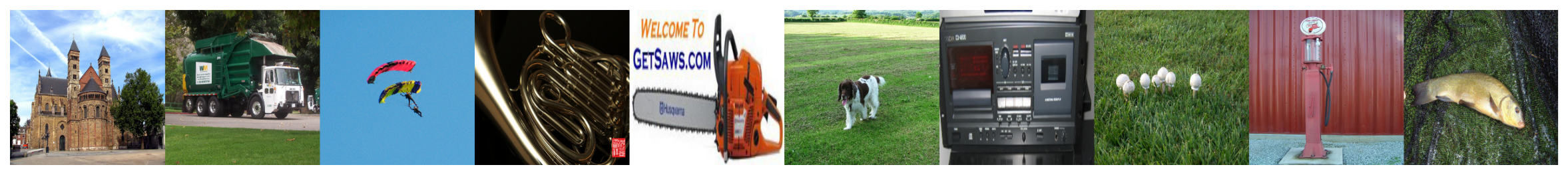}\vspace{-3ex}
  \caption{visualization of the condensed ImageNette.}
  \label{fig:vis_nette}
\end{figure}

The visualizations of our selected images, presented in \cref{fig:vis_1k,fig:vis_tiny,fig:vis_woof,fig:vis_nette}, are randomly sampled from the condensed datasets of ImageNet-1K, Tiny ImageNet, ImageWoof, and ImageNette, respectively.

\end{document}